\pdfoutput=1

\documentclass[11pt]{article}

\usepackage{authblk}
\usepackage[]{EMNLP2022}

\usepackage{times}
\usepackage{latexsym}
\usepackage{graphicx}
\usepackage{amssymb}

\usepackage{physics,amsmath}
\usepackage{cleveref}
\usepackage{enumitem}
\usepackage{booktabs}

\usepackage{textcomp, xspace}
\usepackage{pbox}
\usepackage{array}
\usepackage{multirow}
\usepackage{xcolor}
\usepackage[normalem]{ulem}
\useunder{\uline}{\ul}{}

\crefformat{section}{\S#2#1#3} 
\crefformat{subsection}{\S#2#1#3}
\crefformat{subsubsection}{\S#2#1#3}

\usepackage[T1]{fontenc}

\usepackage[utf8]{inputenc}

\usepackage{microtype}
\usepackage{arydshln} 

\newcommand{\data}{\textsc{SEESAW}\xspace}

\newcommand{\nop}[1]{}

\title{Generative Entity-to-Entity Stance Detection \\with Knowledge Graph Augmentation}

\author[1]{\textbf{Xinliang Frederick Zhang}}
\author[2]{\textbf{Nick Beauchamp}}
\author[1]{\textbf{Lu Wang}}

\affil[1]{Computer Science and Engineering, University of Michigan, Ann Arbor, MI}
\affil[2]{Department of Political Science, Northeastern University, Boston, MA}

\affil[1]{\{\texttt{xlfzhang,wangluxy\}@umich.edu}}
\affil[2]{\texttt{n.beauchamp@northeastern.edu}}

\begin{document}
\maketitle
\begin{abstract}
    Stance detection is typically framed as predicting the sentiment in a given text towards a \textit{target entity}. However, this setup overlooks the importance of the \textit{source entity}, i.e., who is expressing the opinion. 
    In this paper, we emphasize the need for studying interactions among entities when inferring stances. 
    We first introduce a new task, \textbf{entity-to-entity (E2E) stance detection}, which primes models to identify entities in their canonical names and discern stances jointly. 
    To support this study, we curate a new dataset with 10,619 annotations labeled at the \textit{sentence-level} from news articles of different ideological leanings. 
    We present a novel generative framework to allow the generation of canonical names for entities as well as stances among them. We further enhance the model with a graph encoder to summarize entity activities and external knowledge surrounding the entities. 
    Experiments show that our model outperforms strong comparisons by large margins. 
    Further analyses demonstrate the usefulness of E2E stance detection for understanding media quotation and stance landscape, as well as inferring entity ideology. 
\end{abstract}
\section{Introduction}

News media often employ ideological language to sway their readers, including criticizing entities they disagree with, and praising those conforming to their values~\cite{baum2008new,levendusky2013partisan}. 
However, in many cases, the sources do not directly express their sentiments, in part due to the norm that ``objective'' news media should restrict their role to narrating events and quoting others. 
In the realm of political news, many reported events consist of individuals or groups who themselves are engaged in praise or blame. The seemingly neutral act of choosing \textit{who to quote}, and \textit{about what}, as illustrated in Fig.~\ref{tbl:example}, may be shaped by ideology and have significant effects on readers~\cite{gentzkow_shapiro_2006,GENTZKOW2015623}.

\begin{figure}[t]
\small
\centering
\begin{tabular}{ p{72mm} }
\toprule
Trump's rhetoric, including calling Central Americans trying to enter the United States ``an invasion,'' and his hard-line immigration policies have exposed him to condemnation since the El Paso shooting. ``How far is it from Trump's saying this `is an invasion' to the shooter in El Paso declaring `his attack is a response to the Hispanic invasion of Texas?' Not far at all,'' Biden was due to say, according to an advance copy of his speech. {\ul ``In both clear language and in code, this president has fanned the flames of white supremacy in this nation.''} \\ 
\texttt{[0] Joe Biden \colorbox{green!15}{NEG} Donald Trump}\\ 
\texttt{[1] Joe Biden \colorbox{green!15}{NEG} white supremacy}\\
\texttt{[2] Donald Trump \colorbox{magenta!15}{POS} white supremacy} \\
\bottomrule
\end{tabular}
\caption{
Sample stance triplet annotations for a \underline{target sentence}. Entities in \data can be Person or Topic, and are annotated in canonical forms. Multiple stances are expressed, whose inference needs context information, e.g., ``president" refers to Donald Trump.
} 
\vspace{-5mm}
\label{tbl:example}
\end{figure}

There thus exists a pressing need to examine these expressions of support and opposition in news articles~\cite{west2014exploiting} in order to understand how even apparently nonpartisan media can bias readers via the selective inclusion of stances among entities.
Recognizing stances among political entities and events is also important in its own right: if copartisans are more likely to be positive towards each other and vice versa for counter-partisans, this allows us to (1) propagate partisanship and ideology through the signed network~\cite{de2013polarization}, (2) infer ideology not just for politicians, but for events or objects (e.g., a new bill) that may inherently support the positions of specific groups~\cite{diermeier2012language}, and (3) illuminate the implicit ideology of a journalist or media outlet~\cite{hawkins2012motivated}. 

As a first step towards these goals, this paper presents the first study on solving the task of \textbf{entity-to-entity (E2E) stance detection} in an end-to-end fashion: Given a target sentence and its surrounding context, we extract a sequence of stance triplets that can be inferred from the input. Each triplet consists of a \textit{source entity} and their \textit{sentiment} towards a \textit{target entity}, with entities in their canonical forms. 
Existing stance detection methods are largely designed to infer an author's overall sentiment towards a given entity~\cite{sobhani-etal-2017-dataset, li-etal-2021-p} or topic~\cite{VamvasS20, allaway-mckeown-2020-zero} from a text. 
E2E stance detection, by contrast, presents a number of new challenges. 
First, entities can be involved in multiple and even conflicting sentiments within a sentence,\footnote{This may be a common phenomenon due to the journalistic norm of providing ``balanced" views~\cite{MARSHALL200589, Democracy}.} as demonstrated in Fig.~\ref{tbl:example}, suggesting the need to develop a model that can disentangle entity interactions. 
Second, entities are mentioned in various forms, e.g., full names or pronouns. Simply extracting the mentions would cause ambiguity for downstream applications. Canonical names that can be identified via knowledge bases~\cite{Shen2015EntityLW} are thus preferred, which further requires the model to consider contextual information and global knowledge.

In this work, we first collect and annotate an E2E stance dataset, \textbf{\data}\footnote{Our data and code can be accessed at \url{https://github.com/launchnlp/SEESAW}.} (\underline{S}tance between \underline{E}ntity and \underline{E}ntity \underline{S}upplemented with \underline{A}rticle-level vie\underline{W}point), based on $609$ news articles crawled from AllSides.\footnote{\url{https://www.allsides.com}} 
\data contains 10,619 stance triplets annotated at the sentence level, drawn from 203 political news stories, with each ``story'' consisting of 3 articles by media of different ideological leanings, as collected, coded, and aligned by AllSides. Our entities cover people, organizations, events, topics, and other objects. 

We then present a novel encoder-decoder generative framework to output stance triplets in order. 
We first enhance the text encoder with a graph model~\citep{GAT} encoding a semantic graph that summarizes global entity interactions in the context, using relations extracted by an open information extraction (OpenIE) system \citep{Stanovsky2018SupervisedOI}. 
On the decoder side, we improve the transformer decoder block \citep{Vaswani2017AttentionIA} with a task-customized joint attention mechanism to combine textual and graph representations. 
Finally, external knowledge, such as party affiliation or employment, is injected into the graph encoder by adding extra nodes initialized with pretrained representations from Wikipedia. This allows the system to better characterize entity relations. 

We conduct experiments on the newly collected \data to evaluate models' capability of generating stance triplets, and additionally evaluate on a task of stance-only prediction when pairs of entities are given. 
Our model outperforms competitive baselines on E2E stance detection by at least 21\% (relatively, accuracy of 11.32 vs. 13.74), demonstrating the effectiveness of adding knowledge from context and Wikipedia via graph encoding. 
Our best model also outperforms its pipeline counterpart which first extracts entities and then detects sentiment.
This highlights the end-to-end prediction capability of generative models.
Finally, we demonstrate the usefulness of E2E stances for media stance characterization and entity-level ideology prediction. 
Notably, we find that 1) both left- and right-leaning media tend to quote more from the Republican politicians; 
and 2) there appears a \textit{symmetrical asymmetry} in expressed stances: the left is balanced while the right is biased in terms of expressed positivity, but the reverse holds for negativity. 

\section{Related Work}

\subsection{Stance Detection}
Two major types of stance detection are widely studied \citep{Aldayel2021StanceDO}: (1) \textit{sentiment-based}, the goal of which is to uncover the implicit sentiment (favor/against) evinced in texts towards a target, which can be a person or a topic \citep{mohammad-etal-2016-dataset, sobhani-etal-2017-dataset,allaway-mckeown-2020-zero,li-etal-2021-p}; (2) \textit{position-based}, which concerns whether a text snippet agrees/disagrees with a given claim or a headline \citep{ferreira-vlachos-2016-emergent, fakenews,chen-etal-2019-seeing, hanselowski-etal-2019-richly, conforti-etal-2020-will}. In this work, we focus on the \textbf{sentiment-based} stance detection. 
Existing datasets for stance annotations are mainly based on social media posts~\cite{mohammad-etal-2016-dataset, li-etal-2021-p}, making the assumption that the sentiment is always held by the author, thus ignoring source entity annotation. 
Overall, there are at least three limitations for the existing stance detection data: 
1) Data samples are collected within a narrow time period, e.g., a year, about specific events~\cite{sobhani-etal-2017-dataset,li-etal-2021-p}. 
2) Entities are annotated by their mentions only~\cite{deng-wiebe-2015-mpqa, park-etal-2021-blames}, limiting their usage for downstream applications. 
3) Annotations are only available at either sentence-level or article-level, but not both. By contrast, our data spans a 10-year range at both sentence- and article-levels, with entities coded using canonical names. 

Methodologically, existing models are designed for detecting stances towards a specific target entity~\cite{mohammad-etal-2016-dataset,augenstein-etal-2016-stance}. 
However, early methods assume the target entities in test time have been seen during training~\cite{Du2017StanceCW,Zhou2017ConnectingTT,xue-li-2018-aspect}. More recent work uses Large Language Models (LLMs) to enable stance prediction on unseen entities~\citep{li-etal-2021-improving-stance, glandt-etal-2021-stance}. The most similar work to ours are \newcite{zhang-etal-2020-enhancing-cross} and \newcite{liu-etal-2021-enhancing}, both using graph convolutional networks~\cite{GCN} to add external knowledge. Our model is different in at least two key respects:
(1) They use existing knowledge bases, e.g., ConceptNet~\citep{SpeerCH17}, with limited coverage of political knowledge. We instead resort to entity interactions extracted from news articles.
(2) All prior models are based on Transformer encoder \citep{Vaswani2017AttentionIA} only, while we explore the generative power of encoder-decoder models to address the more challenging E2E stance detection task. 
Moreover, none of these methods detects multiple stances from the same input, a gap that this work aims to fill.

\subsection{Generative Models for Classification Task} 
Applying generative models for classification has recently gained research attention, mainly enabled by the wide usage of generative models, especially the large pretrained language models~\cite{brown2020language,raffel2020exploring}.
The most significant advantage of using generative models for classification problems resides in the improved interpretability between label semantics and input samples~\cite{yan-etal-2021-unified, zhang-etal-2021-towards-generative}, especially for multi-label classification problems~\citep{yang-etal-2018-sgm, Liao2020ImprovedSG, Yue2021CliniQG4QAGD}. 
Generative models are especially suitable for our task, since canonical names are preferred in the output.
Recent work~\cite{Humeau2020PolyencodersAA, DeCao2021AutoregressiveER} supports our assumption by showing that
generative models are better at capturing fine-grained interactions between the text and entities than encoder-only models.
This work carries out the first study of deploying such models for a new task, E2E stance detection.
In addition, it extends the model with context information and external knowledge.

\section{\data Collection and Annotation}
\label{data}
\paragraph{Annotation Process.}
We use  AllSides news stories collected by \citet{politics}, where each story contains 1-3 articles on the same event but reported by media of different ideology.
The stories span from 2012 to 2021. 
We only keep news stories with 3 articles and that are pertinent to U.S. politics. 
We manually inspect and select stories to cover diverse topics.
The resulting \data contains 52 topics (full list in Table~\ref{tbl:news_topics}). 
We further clean articles by removing boilerplate and noisy text.

We hired six college students with high English proficiency to conduct annotation tasks. Each person is asked to read all articles written on the same story before annotating. We summarize the main steps below, with detailed protocol in Appendix~\ref{annotation_guideline}. 

\begin{itemize}
\item[1.] They start with reading the article, and then identify entities of political interests that are involved in sentiment expressions. An entity can have a type of person, organization, place, event, topic, or religion. Annotated entities are categorized into \textit{main} and \textit{salient} entities.\footnote{\textit{Main entities} are defined as main event enablers and participants as well as the ones that are affected by such events. \textit{Salient entities} refer to other political or notable figures that appear in the news stories who are not the main entities.}

\item[2.] For \textit{each sentence}, stance is annotated between entities in the triplet format, i.e., <\texttt{source}, \texttt{sentiment}, \texttt{target}> where sentiment can be either \textit{positive} or \textit{negative}.
We use \texttt{Author} as the source entity, if no clear source entity is found.
Also, \texttt{Someone} is used to replace source or target entities of no clear political interest, e.g., ``a neighbor''.

\item[3.] At the article level, we annotate its overall sentiment towards each identified entity, together with the ideology of each entity as well as the ideological leaning of the article, all in 7-way. We then convert the annotations on sentiment and ideological leaning into 3-way and 5-way, respectively.

\end{itemize}

Finally, we conduct cross-document entity resolution by linking annotations to their \textit{canonical names} according to Wikipedia, e.g., ``this president'' becomes ``Donald Trump'' as in Fig.~\ref{tbl:example}.

A quality control procedure is designed to allow annotators to receive timely feedback and improve agreement over time. Details are in Appendix~\ref{data_quality_control}.
Importantly, over 60 randomly sampled articles, the average overlap of annotated entities between pairs of coders is $55.5\%$.
When both source and target entities match, the sentiment agreement level is $97\%$, indicating the high quality of the dataset.

\paragraph{Statistics.} 
\data includes $609$ news articles from $203$ stories, published by 24 different media outlets (9 left, 6 central, and 9 right). Table \ref{tbl:media_outlets} lists all the media outlets. 
On average, each article contains $28$ sentences and $647$ words. 
$44$\% of sentences per article have at least one stance annotation, among which $29$\% have at least two. 

In total, there are $10{,}619$ \textbf{stance annotations} in \data, covering $1,757$ \textbf{distinct entities}. 
$62.4\%$ of the  triplets have \textit{negative} sentiment. 
\textbf{Entity types} in \data cover People ($49.6$\%), Organization ($12.8$\%), Place ($4.2$\%), Event ($12.0$\%), Topic ($17.4$\%), Religion ($0.2$\%), and Others ($3.8$\%), showing its diversity of entity annotations. 
It is worth noting that the \textit{source entity} being \textit{<Author>} and \textit{<Someone>} occurs $9.1\%$ and $12.5\%$ of the annotations. Meanwhile, the number for \textit{target entity} being \textit{<Someone>} is $5.3\%$. 

In terms of \textbf{entity ideology}, the portions of entities annotated as liberal, moderate, and conservative is $31.0\%$, $15.9\%$, and $34.6\%$, respectively.\footnote{Some entities' ideologies are marked as N/A such as Kremlin (non-US entity) and amnesty (topic).} 
Our annotated \textbf{article leanings} align with AllSides' media-level labels for $76.7\%$ of the time.
\section{Generative E2E Stance Detection}
\label{method}

\begin{figure}[t]
    \centering
    \includegraphics[width=0.42\textwidth]{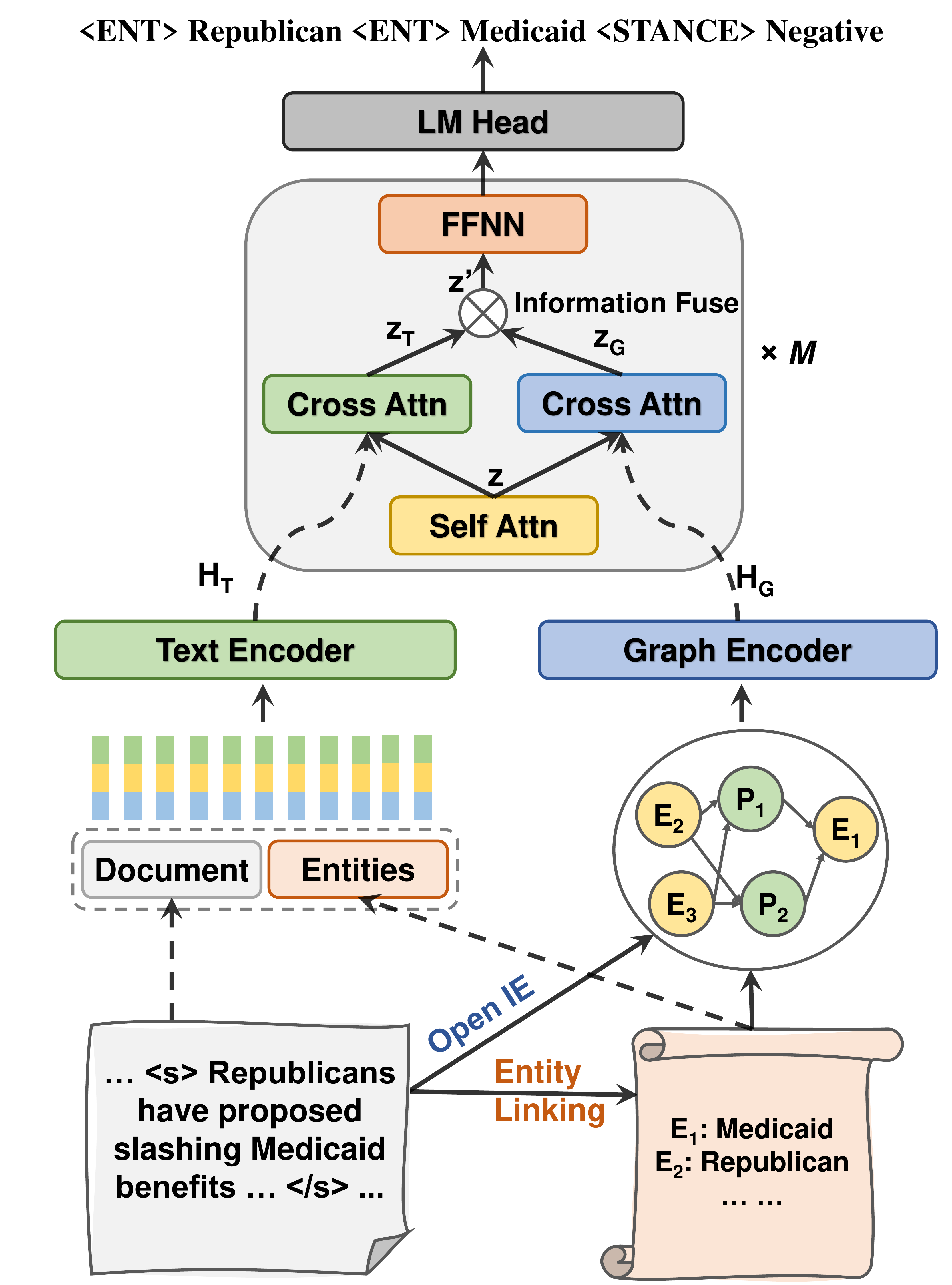}
    \vspace{-2mm}
    \caption{
    Overview of our end-to-end generative framework for stance detection. 
    Our model reads a document $\mathbf{x}$, on which we construct a semantic graph $G$ (\cref{graph_construction}). $G$ contains three types of nodes: entity nodes $E_i$, predicate nodes $P_i$, and Wiki nodes (not shown in the diagram). Extracted entities are paired with document $\mathbf{x}$ and then fed into text encoder, in the format of \textit{``$\mathbf{x}$ <s> <ENT> $E_1$ <ENT> $E_2$ $...$''}. Besides token and position embeddings, we add a third embedding to indicate a token's type: preceding context, target text, succeeding context, or entities.   
    Our decoder implements in-parallel cross-attention (\cref{decoder}) to attend both text ($\mathbf{H}_T$) and node ($\mathbf{H}_G$) representations concurrently. 
    Fused representations are obtained through the information fusion layer.
    }
    \label{fig:architecture}
    \vspace{-5mm}
\end{figure}

Fig.~\ref{fig:architecture} depicts the end-to-end generative framework that reads in a document and produces stance triplets in an auto-regressive fashion, by leveraging multiple knowledge sources using graph augmentation. We use BART~\cite{Lewis2020BARTDS}, a large pretrained encode-decoder model, as the backbone.
Taking as input a \textit{target sentence} from a document $\mathbf{d}$, our model first constructs a semantic graph $G$ (\cref{graph_construction}), which is encoded via the graph encoder (\cref{graph_encoder}). 
The contextualized representations of tokens and nodes are denoted as $\mathbf{H}_T$ and $\mathbf{H}_G$. 
Stance triplets are generated by our decoder using improved \textit{in-parallel attention} and \textit{information fusion} mechanisms (\cref{decoder}). 
Moreover, we inject Wikipedia knowledge to support the identification of relations between entities (\cref{wiki}). 
Below we describe the main components, with additional formulation and implementation details in Appendix~\ref{model_details}.

\subsection{Local Semantic Graph Construction}
\label{graph_construction}
Our goal is to construct a semantic graph that can summarize events and sentiments involving entities in the document context. We thus use OpenIE \citep{Stanovsky2018SupervisedOI} to obtain semantic relation outputs in the form of <subject, predicate, object>. Triplets whose span is longer than 15 tokens are dropped. We also conduct global entity linking\footnote{\url{https://cloud.google.com/natural-language/}} to extract canonical names for all entities in the document, which are fed into the text encoder as shown in Fig.~\ref{fig:architecture}.
This linking step facilitates the generation of canonical names in a consistent manner.

We start constructing the graph $G$ by treating the extracted entities as \textit{entity nodes}, where co-referential mentions of the same entity are collapsed into a single node. 
Following \citet{Beck2018GraphtoSequenceLU} and \citet{huang-etal-2020-knowledge}, we further create \textit{predicate nodes}.
We then add directed edges from subject to predicate and from predicate to object.
We  add reverse edges and self-loops to enhance graph connectivity and improve information flow.

\subsection{Graph Encoder}
\label{graph_encoder}

We {initialize} node representations $\mathbf{H}_G$
by using the average contextual token embeddings ($\mathbf{H}_T$) of all co-referential mentions.
Similar to \citet{yasunaga-etal-2021-qa}, we add a \textit{global} node, initialized with mean pooling over all tokens in the target sentence. The global node is connected to entity nodes in $G$ to allow better communication of text knowledge.

Our graph encoder improves upon Graph Attention Networks \citep[GAT;][]{GAT} using Transformer layer networks and \textit{Add \& Norm} structures \citep{Vaswani2017AttentionIA}. 
Concretely, in each layer, we use the multi-head GAT massage passing rule to update node representations $\mathbf{H}_G$, and then pass them through a 2-layer MLP.
Residual connections \citep{He2016DeepRL} and layer normalization \citep{Ba2016LayerN} are employed to stabilize the hidden state dynamics.
We use two layers with 8-head GAT to produce final node representations $\mathbf{H}_G$.

\subsection{Decoder}
\label{decoder}
We further improve the Transformer decoder to enable reasoning over both the text and the graph. 
The key difference between the vanilla Transformer decoder and ours is the \textit{in-parallel cross-attention} layer which allows better integration of knowledge encoded in the two sources. 
Concretely, in-parallel cross attentions are implemented as follows: 
\vspace{-1.5mm}
\begin{equation} 
\begin{aligned}
\mathbf{z}_T = & \textrm{LayerNorm}(\mathbf{z}+\textrm{Attn}(\mathbf{z},\mathbf{H}_T))\\
\mathbf{z}_G = & \textrm{LayerNorm}(\mathbf{z}+\textrm{Attn}(\mathbf{z},\mathbf{H}_G))
\end{aligned}
\vspace{-1.5mm}
\end{equation}
\noindent where $\mathbf{z}$ denotes the output from the self-attention layer and  Attn($\cdot$,$\cdot$) is the same cross-attention mechanism as implemented in \citet{Vaswani2017AttentionIA}.  

Next, we introduce an \textit{information fusion} module to enable the interaction between textual ($\mathbf{z}_T$) and graph ($\mathbf{z}_G$) hidden states, in order to obtain the fused representation, $\mathbf{z'}$.
We implement two information fusion operations: (1) \textbf{addition}, i.e., $\mathbf{z'} =  \mathbf{z}_T + \mathbf{z}_G$, and (2) \textbf{gating} mechanism between $\mathbf{z}_T$ and  $\mathbf{z}_G$ as in \citet{ZhaoNDK18} except that we use GELU($\cdot$) as the activation function. The operation selection  is determined by  downstream tasks.

\subsection{Training Objectives}
\label{training}
We adopt the cross entropy training objective for model training, $\mathcal{L}_{stance}$. The reference $\mathbf{y}$ is a sequence of ground-truth stance triplet(s), sorted by their entities' first occurrences in the target sentence. Input and output formats are shown in Fig.~\ref{fig:architecture}.

\smallskip
\noindent\textbf{Variant with Node Prediction.} 
In addition to modeling entity interactions in the graph, we enhance the model by adding an auxiliary objective to predict node salience, i.e., whether the corresponding entity should appear in the stance triplets $\mathbf{y}$ to be generated. 
This is motivated by the observation that $G$ usually contains excessive entity nodes, only a small number of which are involved in sentiment expression in the target sentence. 
Specifically, for each entity node, we predict its salience by applying affine transformation over its representation $\mathbf{h}_G$, followed by a sigmoid function to get a single value in $[0,1]$. We adopt the binary cross entropy (BCE) objective to minimize the loss $\mathcal{L}_{node}$ over all \textit{entity nodes}. 
Finally, when the node prediction module is enabled, the overall loss for the \textbf{multitask} learning setup is $\mathcal{L}_{multi} = \mathcal{L}_{stance} + \mathcal{L}_{node}$.

\subsection{Graph Expansion: Wiki Knowledge Injection}
\label{wiki}
To gain a better understanding of stances among entities over controversial issues, it is useful to access external knowledge about the entities mentioned in the news, e.g., their party affiliations. 
Therefore, we obtain the knowledge representations for entities using Wikipedia2Vec~\cite{wiki2vec}, a tool that jointly learns word and entity embeddings from Wikipedia.
The learned entity embeddings are shown to be effective in encoding the background knowledge discussed in Wikipedia. 
These embeddings are then added as \textit{wiki nodes} in graph $G$. 
We add edges between an entity node and a wiki node, if the entity is linked to the corresponding Wikipedia entry based on entity linking.\footnote{\url{https://cloud.google.com/natural-language/}} 
In our implementation, we take the pre-trained 500-dimensional vectors,\footnote{\url{wikipedia2vec.s3.amazonaws.com/models/en/2018-04-20/enwiki_20180420_500d.pkl.bz2}} transformed by a two-layer MLP, for node representations initialization.

To summarize, graph-augmented generative models have been studied for several generation tasks, including abstractive summarization~\cite{huang-etal-2020-knowledge}, question answering~\cite{yasunaga-etal-2021-qa}, and question generation~\cite{su-etal-2020-multi}. 
Prior design of graph encoders uses either external knowledge bases~\cite{greaseLM} or a local graph constructed using semantic parsing~\cite{cao-wang-2021-controllable}.
Since large-scale structured knowledge base does not exist for the political domain, our method differs from previous work in that we leverage both entity interactions from the context and external knowledge from Wikipedia in a unified graph representation learning framework to better characterize entity interactions.

\section{Experiments}
\label{experiemnts}

\begin{table}[t]
\centering
\resizebox{0.99\linewidth}{!}{%
\begin{tabular}{lrr}
\toprule
 & \data (Task A) & \citeauthor{park-etal-2021-blames} (Task B) \\
 \midrule
Target Sentence length & 30.3 & 31.0 \\
Label ratio (pos/neg) & 37.6\%/62.4\% & 35.3\%/64.7\% \\
Splits (train/valid/test) & 4505/1313/1378 & 4252/506/562 \\
\bottomrule

\end{tabular}
}
\caption{Statistics of the two datasets for experiments. Data by \citet{park-etal-2021-blames} only contains single sentences without context.
We split the \data chronologically, and use the same splits as in \citet{park-etal-2021-blames}.}
\label{tbl:data_stat}
\vspace{-5mm}
\end{table}

\subsection{Tasks and Datasets}

We conduct evaluation on two different stance detection tasks. Table~\ref{tbl:data_stat} shows the basic statistics of datasets and splits.

\smallskip
\noindent\textbf{Task A: Entity-to-entity stance detection.} 
We experiment with \data for generating stance triplets of <source, sentiment, target>. 
The input text can be a target sentence alone or with surrounding context (up to $k$ preceding and $k$ succeeding sentences). We set $k=3$ for all experiments. 

\smallskip
\noindent\textbf{Task B: Stance-only prediction for pairwise entities.} 
\citet{park-etal-2021-blames} build a dataset annotating sentiments between mentions rather than canonical entities. We include this dataset to assess our model's capability on a stance-related classification task. For experiments, we only keep samples with positive and negative sentiments. 
Formally, their input contains one sentence $s$ and two entities $e_1$ and $e_2$. The goal is to jointly predict the direction and the sentiment, i.e., four labels in total. 

\subsection{Baselines}
For Task A, since there is no existing E2E stance detection models, we consider finetuning BART using different inputs as baselines: 
(1) \textbf{sentence}: target sentence only; (2) \textbf{sentence + context}: target sentence with surrounding context; (3) \textbf{sentence + context + entities}: additionally appending all entities in their canonical names as extracted in \cref{graph_construction}, same as our model's input. 

We further consider two variants of our model as baselines. We first design a \textbf{pipeline model}, which first uses the node prediction module to identify salient entities for inclusion in the stance triplets. Then we introduce a soft mask layer over entity nodes in $G$ before passing them into the graph encoder, by multiplying node representations with their predicted salience scores. We also experiment with \textbf{oracle entities}, where we feed in the ground-truth salient entities for text encoding and graph representation learning. 

Finally, to test the effectiveness of our designed in-parallel cross-attention, we compare with a \textbf{sequential attention}, designed by \citet{cao-wang-2021-controllable}, to consolidate text and graph modalities. They allow the decoder hidden states to first attend token and then node representations. Their model differs from our model only in the attention design.

On Task B, since LLMs have obtained the state-of-the-art performance on existing stance prediction tasks~\cite{glandt-etal-2021-stance, politics}, we compare with the following LLM-based methods in addition to \textbf{BART}. 
We compare with \textbf{DSE2QA}~\citep{park-etal-2021-blames}, which is built on top of RoBERTa \citep{liu2019roberta}. They transform the sentiment classification task into yes/no question answering, where the questions ask whether a sentiment can be entailed from several manually designed questions appended after the target sentence. We use the question that obtained the best performance on their dataset, i.e., ``$e_1$ - $e_2$ - [sentiment]''. 
We then consider a recent LLM, \textbf{POLITICS} \citep{politics}, trained on RoBERTa with ideology-driven pretraining objectives that compare articles on the same story.

\subsection{Evaluation Metrics}

For both tasks, we report accuracy and F1 scores. For Task A, accuracy is measured at the sample level, i.e., all stance triplets need to be generated correctly to be considered as correct. 
F1 is instead measured at the triplet level. 
We include another metric, accuracy-\textbf{any}, where for each sample, the prediction is considered correct if at least one triplet is found in the reference. 
We also break down the triplets, and evaluate varying \textbf{aspects} based on pairs of source entity-sentiment (\textbf{src-s}), sentiment-target entity (\textbf{s-tgt}), and source-target entities (\textbf{src-tgt}), using accuracy-{any}.

\subsection{Results}
 \label{main_results}

\begin{table}[t]
\centering
\resizebox{1.0\linewidth}{!}{%
\begin{tabular}{lcccccc}
\toprule
\multicolumn{1}{c}{\multirow{2}{*}{}} & \multicolumn{3}{c}{\textbf{Full}} & \multicolumn{3}{c}{\textbf{Aspect}} \\ 
\cmidrule(lr){2-4}
\cmidrule(lr){5-7}
 & Acc. & F1 & Acc.Any & src-s & s-tgt & src-tgt \\
 \midrule
 \multicolumn{7}{l}{\textbf{Baselines (no graph)}}\\
Sentence (Sen) & 7.26 & 10.35 & 12.39 & 36.66 & 23.81 & 14.88 \\
Sen + Context (Ctx) & 9.66 & 14.08 & 16.87 & 45.24 & 27.79 & 20.01 \\
Sen + Ctx + Entities & 11.32 & 16.00 & 19.15 & 47.43 & 30.03 & 23.18 \\

\midrule
\multicolumn{7}{l}{\textbf{Pipeline Models (Ours)}}\\
Graph & 12.03 & 15.77 & 19.86 & 46.60 & 31.07 & 23.19 \\
\noalign{\vskip 0.3ex}
\hdashline
\noalign{\vskip 0.3ex}
\,\, +  Oracle Entities & 31.84 & 35.58 & 44.87 & 66.33 & 55.50 & 53.82 \\
\midrule
\multicolumn{7}{l}{\textbf{End-to-end Models (Ours)}}\\
Graph (seq. attn.) & 12.97 & 17.22 & 20.62 & 50.58 & 32.45 & 24.76 \\
Graph  &  13.62 & 18.12 & 21.78 & 51.01 & 32.65 & \textbf{26.08} \\
\,\, + Multitask & 13.34 & 18.16 & 21.77 & \textbf{52.10} & 32.06 & 26.07 \\
\,\, + Wiki & \textbf{13.74} & \textbf{18.24} & \textbf{21.87} & 51.41 & \textbf{32.69} & 25.94\\
\bottomrule
\end{tabular}
}
\vspace{-2mm}
\caption{
Results on \data for E2E stance detection task, and breakdown of accuracy scores by aspects (average of 5 runs). 
Best results without oracle entities are in \textbf{bold}. 
Our graph-augmented model with Wikipedia knowledge performs the best on 4 out of 6 metrics, indicating the effectiveness of encoding knowledge. 
Results with standard deviation are in Table~\ref{tbl:result_std}. 
}
\label{tbl:result}
\vspace{-2mm}
\end{table}

Results for E2E stance detection is displayed in Table~\ref{tbl:result}. 
Compared with baselines, we see significant improvements by providing context and entities in canonical forms, indicating that  adding story context and additional knowledge about entity names is useful for the E2E stance detection task. 

Next, though the pipeline variant of our model provides better explainability as it first identifies salient entities, it yields inferior performance than the end-to-end version of our model. After inspection, we find that the salient entity prediction module only reaches around $58\%$ for F1.
With the oracle entities as input, we see a significant boost in the performance, highlighting the importance and difficulty of entity understanding and extraction. 

Importantly, our model enhanced with Wikipedia knowledge performs the best on 4 out of 6 metrics. 
This signifies the effective design of graph modeling on entity interactions.  
Moreover, our newly designed in-parallel attention is also shown to be more effective than attending the two sources of information in sequence, as done in~\citet{cao-wang-2021-controllable}. This implies that having symmetric integration of text and graph can be important, though this should be tested on other tasks in future work. 
When breaking down the predicted stance triplets into different pairs, we see that identifying source entity and sentiment is easier than predicting sentiment and target entity.
This might be because target entities are often introduced earlier in the article, thus requiring long-term discourse understanding.

Finally, the overall performance of E2E stance detection is quite low for all models. This is mainly because  models may fail to generate exactly the same canonical names for entities and often fall short of producing all stance instances when multiple sentiments are embedded in a single sentence. 

\begin{table}[t]
\centering
\resizebox{0.9\linewidth}{!}{%
\begin{tabular}{lrr}
\toprule
 & Accuracy & Macro F1 \\
\midrule
BART \citep{Lewis2020BARTDS} & 86.32 & 77.53 \\
POLITICS \citep{politics} & 86.33 & 77.48 \\
DSE2QA \citep{park-etal-2021-blames} & 87.78 & \textbf{79.90} \\
\midrule
Our Model & \textbf{87.79} & 79.01 \\
\bottomrule
\end{tabular}
}
\vspace{-2mm}
\caption{
Results on stance-only prediction for specified pairwise entities.
Our model performs on par with state-of-the-art models in stance detection tasks (POLITICS and DSE2QA). Results with std. deviation in Table~\ref{tbl:result_dse_std}.}
\label{tbl:result_dse}
\vspace{-3mm}
\end{table}

On the stance-only prediction task, Table~\ref{tbl:result_dse} shows that our model yields better or comparable performance than the state-of-the-art models. 
This demonstrates that our generative stance detection model can also perform well on a quaternary classification setup. Note that the input text is short ($\sim$30 tokens), limiting our model's capability of capturing global context. However, our model still outperforms BART and the recently trained LLM, POLITICS, designed for ideology prediction and stance detection. 
The experimental results are inline with \citet{Lewis2020BARTDS} that the improvements on generation tasks do not come at the expense of classification performance.

\begin{figure}[t]
\small
\centering
\begin{tabular}{ p{70mm} }
\toprule
Ms. Harris, who is making her first trip to a battleground state since joining the Democratic ticket, is visiting with union workers and leaders as well as African-American businesspeople and pastors in Milwaukee, the Black hub of the state. {\ul Each is expected to focus on the economy, with Mr. Pence hailing \textbf{the state}'s job growth before the coronavirus pandemic and Ms. Harris critiquing the administration's handling of the virus and the resultant impact on the economy.} Yet their political missions are different. The vice president is hoping to appeal to voters in a historically Democratic part of \textbf{Wisconsin}, where Mr. Trump outperformed his Republican predecessors, in hopes they abandon their political roots again. \\ 
\texttt{[0] Mike Pence  \colorbox{magenta!15}{POS} Wisconsin}\\ 
\texttt{[1] Kamala Harris \colorbox{green!15}{NEG} Donald Trump}\\
\texttt{[2] Kamala Harris \colorbox{green!15}{NEG}  $<$Someone$>$ } \\
\noalign{\vskip 0.3ex}
\hdashline[5pt/4pt]
\noalign{\vskip 0.3ex}
\textbf{Sent.}: [0] Mike Pence \colorbox{magenta!15}{POS} $<$Someone$>$  \\
\textbf{Sent. + Cxt.}: [0] Mike Pence \colorbox{magenta!15}{POS} $<$Someone$>$  \\
\textbf{Sent. + Cxt. + Ent.}: [0] Mike Pence \colorbox{magenta!15}{POS} $<$Someone$>$    \\
\textbf{Graph model} (ours):  [0] Mike Pence \colorbox{magenta!15}{POS} job growth; [1] Kamala Harris \colorbox{green!15}{NEG} Donald Trump \\
\textbf{Graph model + Wiki} (ours): [0]  Mike Pence  \colorbox{magenta!15}{POS} Wisconsin; [1] Kamala Harris \colorbox{green!15}{NEG} Donald Trump \\
\bottomrule
\end{tabular}
\vspace{-2mm}
\caption{
Sample system predictions (below the dotted line) with human labeled triples (above the dotted line). Target sentence is \underline{underlined}. 
All three baselines fail to identify the correct target entity. 
By contrast, our graph-augmented end-to-end 
model predicts the first triplet by leveraging the semantic relation as captured by graph $G$. 
After encoding Wikipedia knowledge, our model draws the connection between Wisconsin and ``the state'' in the text, thus generating a correct stance triplet [0].  
Our models also  produce multiple triplets. 
} 
\vspace{-2mm}
\label{tbl:error_analysis}
\end{figure}

\subsection{Sample Output and Error Analysis}
\label{additional_error_analysis}

Fig.~\ref{tbl:error_analysis} shows one test example with system outputs from baselines and our models. 
All three baseline models detect a positive sentiment ascribed to Mike Pence but fail to uncover the specific target entity. However, our graph-augmented model manages to produce stance triplet $[1]$, using the direct edge linking Kamala Harris and Donald Trump through a negative predicate in the corresponding graph $G$. 
We also find that our model using Wikipedia knowledge often uncovers hidden relations between entities, e.g., party affiliations and geopolitical relations, which are useful for stance detection but cannot be inferred from the document  alone. Such as in this example, to enable the generation of stance triplet $[0]$, our model leverages Wikipedia knowledge to draw the connection between Wisconsin and ``\textit{the state}'' in the news. 
Moreover, this example showcases the power of our model in generating multiple stances, which is an essential capability for the E2E stance detection task. In Fig.~\ref{tbl:error_analysis_additional}, we show another example, where none of the models produces a correct stance triplet, further confirming the challenge posed by E2E stance detection. This points out the future direction of investigating more powerful models that can better make inferences and perform reasoning over knowledge.

\section{Further Analysis}

\begin{figure}[t]
    \centering
    \includegraphics[width=0.42\textwidth]{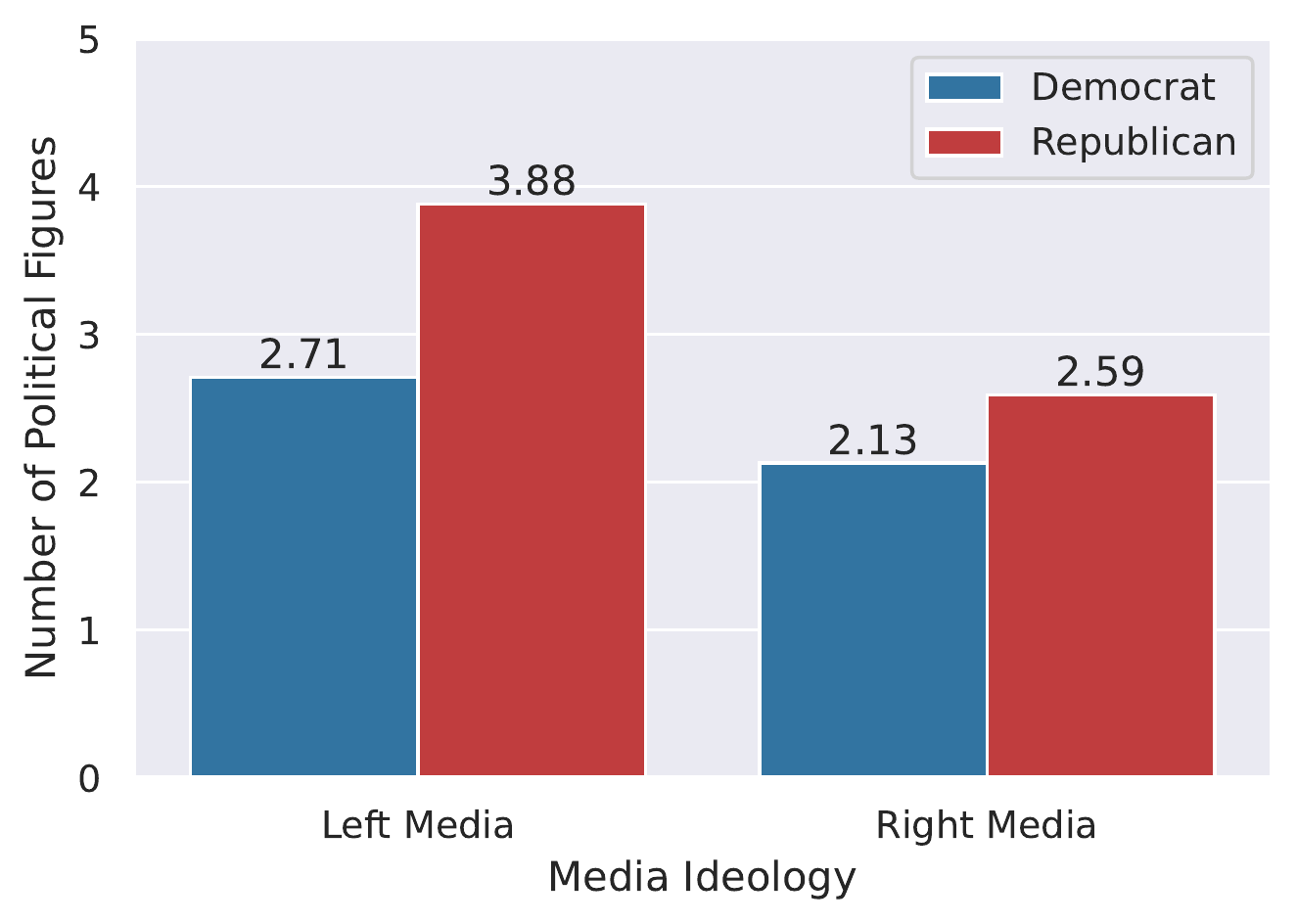}
    \vspace{-3mm}
    \caption{Media quoting Democrats vs. Republicans by counting source entities per article. Both left- and right-leaning media outlets quote Republicans more. 
    }
    \label{fig:analysis_0}
    \vspace{-2mm}
\end{figure}

In \data, we are able to identify the partisanship of 204 politicians (Democrat vs. Republican) based on Voteview.\footnote{\url{https://voteview.com/}} This subset accounts for more than 60\% of person mentions in the dataset. We further include \textit{Democrat} and \textit{Republican} as two separate entities, since they are also frequently mentioned in news. Analyses in this section are done based on this entity set (henceforth \textit{analysis set}).

\subsection{Landscape of Media Quotation and Stance}
We start with examining the relation between media ideology and their stances. 
We first study \textit{do media tend to quote people of the same or opposite ideologies?} 
To answer this question, we count the average number of political figures quoted as source entities in each article. 
Interestingly, media from both sides are more likely to quote Republican politicians, as seen in Fig.~\ref{fig:analysis_0}. 
This is consistent with recent study on U.S. TV media~\cite{Hong2021AnalysisOF}, where the authors show that Republicans receive more screen time as well as get longer TV interviews time than Democrats.\footnote{By original authors~\cite{Hong2021AnalysisOF}, the conclusion of interview time might not be true due to biased data sampling.} 
Additionally, left-leaning outlets use more quotes than their right counterparts, which also aligns with previous observations~\cite{welch1998state}. 

\begin{figure}[t]
    \centering
    \includegraphics[width=0.42\textwidth]{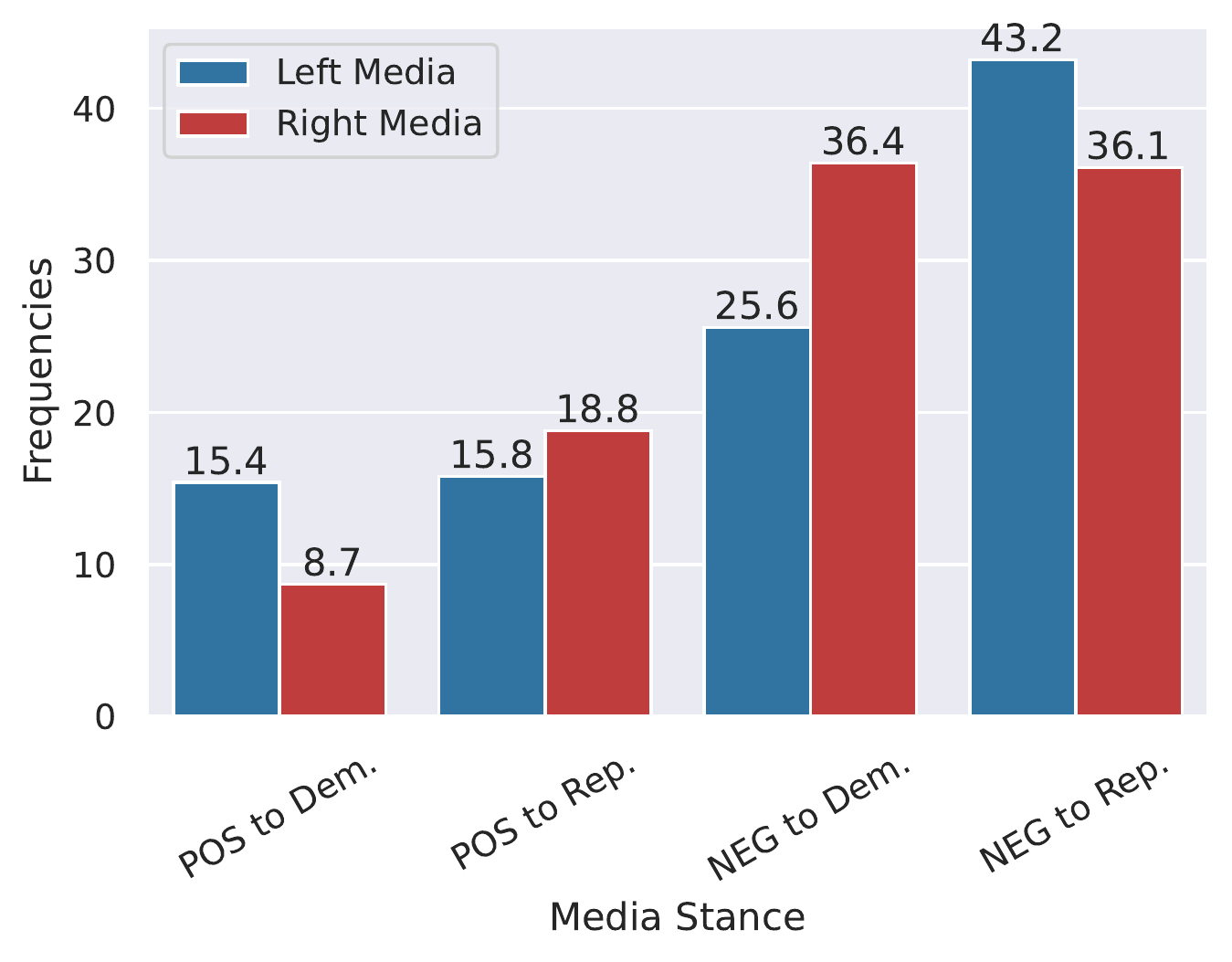}
    \vspace{-4mm}
    \caption{
    Percentage of stance triplets that media favoring or criticizing entities from the same or the opposite side. Media of both sides attack politicians from the opposite parties more than their own parties. 
    Note there is a \textit{symmetrical asymmetry} phenomenon: Left is balanced while the right is unbalanced in terms of indicated positivity, and the other way around for  negativity. 
    }
    \label{fig:analysis_0b_p}
    \vspace{-2mm}
\end{figure}

Next, we ask: \textit{do media tend to be more positive towards people with the same ideology as theirs and be more negative towards out-group entities?} 
Here we consider stance triplets containing the target entities in the analysis set. 
First, as can be seen from Fig.~\ref{fig:analysis_0b_p}, 
more negative sentiments are observed in news articles, which align with existing work that shows journalists more often report blame than praise~\cite{edz054}. 
More importantly, we observe an interesting sentiment pattern of \textbf{symmetrical asymmetry}:  Left-leaning media produces articles use similar amounts of positivity towards Democrats and Republicans ($15.4\%$ vs. $15.8\%$), while right-leaning media are more positive towards Republicans ($18.8\%$ vs. $8.7\%$). By contrast, when it comes to negativity, right-leaning media are more balanced ($36.4\%$ vs. $36.1\%$), while left-leaning media are unbalanced ($25.6\%$ to Democrat vs. $43.2\%$ to Republicans). This suggests that the left and right media may be biased in different ways: the left by directing more negativity to the opposing side, the right by directing more positivity towards their own side.

\subsection{E2E Stances for Ideology Prediction} 
Here we test whether the knowledge of E2E stances can help with entity-level ideology prediction. 
Based on the sentiments expressed among politicians, we construct a directed graph with edges indicating the direction and sentiment between entities. 
We random mask the ideology of $k$\% of the nodes, and then infer their ideology using sentiments from/to their neighbors.
Each node $e$ has two counters: $c^{D}$ for Democrat and $c^{R}$ for Republican.
For each of $e$'s neighbors with known ideology, $c^{D}$ increases by $1$ if (1) the neighbor is {D} and positive sentiment is expressed in either direction, or (2) the neighbor is {R} and negative sentiment is observed in either direction.
Similar rules are used for $c^{R}$.
The counter with the higher value decides $e$'s ideology. 

\begin{figure}[t]
    \centering
    \includegraphics[width=0.41\textwidth]{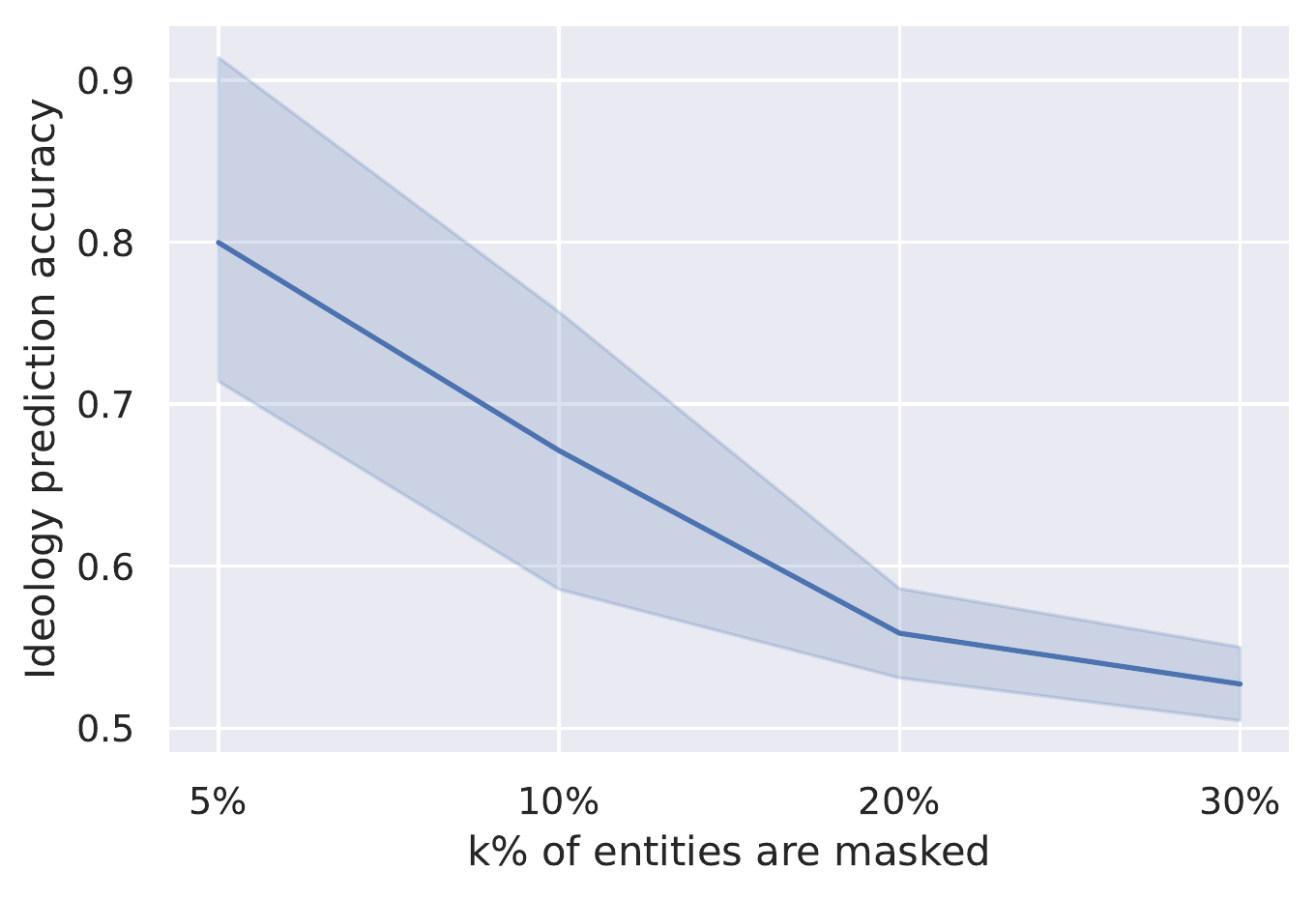}
    \vspace{-4mm}
    \caption{
    Entity-level ideology prediction using stances from/to their neighboring entities with known ideology. We increase the ratio of entities being masked, which decreases the ideology prediction accuracy. This implies  knowing entity's support/oppose interactions with other entities is helpful for predicting their own ideology. 
    }
    \label{fig:impact_masking}
    \vspace{-2mm}
\end{figure}

By varying the percentage of nodes to be masked (Fig.~\ref{fig:impact_masking}), we observe that, the more we know about an entity's sentiment interactions with others, the more accurate we can predict their ideology.
This shows the usefulness of networks constructed from E2E stances as inferred from news articles.

\subsection{Inter- and Intra-group Sentiment} 
Finally, we observe that the majority of inter-party stances are negative, e.g., $92.7\%$ of sentiment by Democratic politicians towards Republicans is negative, and the number of republicans is $91.9\%$. This is unsurprising given the current level of polarization in the U.S. \citep{polarization1, polarization2}. Notably, the Republican Party is much more divided compared with Democrats, where more than half of intra-group stances (i.e., $56.0\%$) within Republican Party carry negative sentiments, whereas the percentage for Democrats is only $25.2\%$. These results contradict recent observations for in-group sentiment as measured on social media users and congressional members~\cite{grossmann2016asymmetric,benkler2018network}. This highlights the significance of studying stances quoted by news media and suggests new avenues for future research.

\section{Conclusion}

We present and investigate a novel task: entity-to-entity (E2E) stance detection, with the goal of extracting a sequence of stance triplets from a target sentence by jointly identifying entities in their canonical names and discerning stances among them. To support this study, we annotate a new dataset, \data, with 10,619 sentence-level annotations. 
We propose a novel end-to-end generative framework to output stance triplets. Specifically, we enhance standard encoder-decoder models with a semantic graph to capture entity interactions within context.
We further augment our model with external knowledge learned from Wikipedia, yielding the best overall performance.
We conduct further analyses to demonstrate the effectiveness of E2E stances on media landscape characterization and entity ideology prediction.
\section*{Acknowledgments}
This work is supported in part through National Science Foundation under grant IIS-2127747, Air Force Office of Scientific Research under grant FA9550-22-1-0099, and computational resources and services provided by Advanced Research Computing (ARC), a division of Information and Technology Services (ITS) at the University of Michigan, Ann Arbor. 
We appreciate the anonymous reviewers for their helpful comments. 
We thank David Wegsman, Isabella Allada, Margaret Peterson, Jiayi Zhang and Bingyan Hu for their efforts in SEESAW construction. We also thank Jiayi Zhang and Hongyi Yin for conducting data  checking.
\section*{Limitation}

\subsection*{GPU resources}

The framework proposed in this work is an encoder-decoder based generative model.
It is thus more time-consuming than standard discriminative models for training and evaluation, which in turn results in higher carbon footprint.
Specifically, we run our experiments on 1 single NVIDIA RTX A6000 with significant CPU resources.
The training time for our model is usually around 5 hours.

\subsection*{System limitation}

In spite of achieving the best performance on E2E stance detection and comparable performance with the SOTA model \citep{park-etal-2021-blames} on the task of stance-only prediction given pairwise entities, our model is still limited in the following aspects. (1) From Table~\ref{tbl:result}, even the best model struggles with extracting correct <source, target> pairs. (2) Though we have pre-processed the data and conducted global entity linking, which helps with entity-level coreference resolution, better designs are needed to help resolve event coreference. Concretely, as shown in Fig.~\ref{tbl:error_analysis_additional}, our best model still suffers from making sense of the correct relation between ``such an action'' and ``affidavit''.

\subsection*{Evaluation limitation}

We believe the high-quality annotations and diverse entities in our \data can help foster research along this novel research direction. However, the adopted evaluation schemes still have their own shortfalls.
For example, in Fig.~\ref{tbl:error_analysis}, our model's output, \textit{Mike Pence \colorbox{magenta!15}{POS} job growth}, can be considered as correct. 
Yet, under the current automatic evaluation scheme, this prediction is counted as a mistake.
More robust and accurate metrics need to be developed to gauge the research progress.

\section*{Ethical Consideration}

\subsection*{\data}

\noindent \textbf{\data collection.} All news articles were collected in a manner consistent with the terms of use of the original sources as well as the intellectual property and the privacy rights of the original authors of the texts, i.e., source owners. During data collection, the authors honored privacy rights of content creators, thus did not collect any sensitive information that can reveal their identities. All participants involved in the process have completed human subjects research training at their affiliated institutions. We also consulted Section 107\footnote{\url{https://www.copyright.gov/title17/92chap1.html\#107}.} of the U.S. Copyright Act and ensured that our collection action fell under the fair use category.

\smallskip
\noindent\textbf{\data annotation.} In this study, manual work is involved. All the participants are college students, who participated in the this project for credits rather than compensation. We treat every annotator fairly by holding weekly meetings to give them timely feedbacks and grade them quite leniently to express our appreciation for their consistent efforts.

\smallskip

\subsection*{Benefit and Potential Misuse of our developed Systems and \data}

\noindent \textbf{Intended use.} The models developed in this work can assist the general public to measure and understand stance evinced in texts. For example, our model can be deployed in wild environments to automatically extract stance triplets at no cost.

\smallskip
\noindent \textbf{Failure mode} is defined as situations where our model fails to correctly extract a stance triplet  of a given text. In such cases, our model might deliver misinformation or cause misunderstanding towards a political figure or a policy. For vulnerable populations (e.g., people who maybe not be able to make the right judgements), the harm could be tremendously magnified when they fail to interpret the model outputs or blindly trust systems' outputs. Ideally, the interpretation of our model's predictions should be carried out within the broader context of the source text.

\smallskip
\noindent \textbf{Misuse potential.}  Users may mistakenly take the machine prediction as a golden rule or a fact. We would recommend any politics-related machine learning models, including ours, put up an ``use with caution'' message to encourage users to check more sources or consult political science experts to reduce the risk of being misled by single source. 
Moreover, our developed system might be misused to label people with a specific stance towards an issue that they do not want to be associated with. We suggest that when in use the tools should be accompanied with descriptions about their limitations and imperfect performance, as well as allow users to opt out from being the subjects of measurement.

\smallskip
\noindent \textbf{Biases and bias mitigation.} No known bias is observed in \data since we collected balanced views of news stories from AllSides. During annotations, annotators were not biased since they have the full access to all articles reporting on the same event but published by media of different ideology. Meanwhile, our developed systems were not designed to encode bias. In the training phase, we split the data on the story level, i.e., one story consisting of three articles from different ideologies, and we believe such training paradigm would help mitigate bias to a certain degree. 

\smallskip
\noindent \textbf{Potential limitation.} Although balanced views are considered, the topic coverage in \data is not exhaustive, and does not include other trending media or content of different modalities for expressing opinions, such as TV transcripts, images, and videos. Thus, the predictive performance of our developed system may still be under investigated.

In conclusion, there is no greater than minimal risk/harm introduced by either our dataset \data or our developed novel system using it. However, to discourage misuse of \data or  stance detection related systems, we will always warn users that systems' outputs are for informational purpose only and users should always resort to the broader context to reduce the risk of absorbing biased information.

\clearpage

\clearpage
\setcounter{table}{0}
\setcounter{figure}{0}
\renewcommand{\thefigure}{A\arabic{figure}}
\renewcommand{\thetable}{A\arabic{table}}

\appendix

\begin{table}[]
\centering
\resizebox{0.8\linewidth}{!}{%

\begin{tabular}{lr}
\toprule
  News topic & \# of news stories \\ 
\midrule
Elections & 30\\
Politics & 16\\
White House & 15\\
US House & 11\\
US Senate & 10\\
Immigration & 10\\
Violence in America & 7\\
Federal Budget & 7\\
Gun Control and Gun Rights & 7\\
Healthcare & 6 \\
US Congress & 5\\
Coronavirus & 5\\
Supreme Court & 4 \\
Justice Department& 4\\
National Security& 4\\
National Defense& 4\\
State Department& 3\\
Economic Policy& 3\\
Terrorism & 3\\
Economy and Jobs & 3\\
LGBT Rights & 3\\
Labor & 2\\
Holidays & 2\\
Race and Racism & 2\\
Nuclear Weapons & 2 \\
FBI & 2\\
Justice& 2\\
Sexual Misconduct& 2\\
Abortion& 2\\
Education& 2\\
Impeachment & 2\\
Free Speech& 2\\
Treasury& 2 \\
Republican Party&1\\
Religion and Faith& 1\\
Fake news& 1\\
Campaign Finance & 1\\
Inequality & 1\\
Donald Trump &  1\\
Homeland Security & 1\\
US Military& 1 \\
Public Health& 1\\
Criminal Justice & 1\\
Voting Rights and Voter Fraud & 1\\
Joe Biden & 1\\
NSA & 1\\
Veterans Affairs & 1\\
Cybersecurity & 1\\
World & 1\\
Middle East & 1\\
Family and Marriage & 1\\
Taxes & 1\\
\midrule
Total & 203 \\
\bottomrule
\end{tabular}
}
\caption{Number of news stories for each topic in \data.}
\vspace{-15pt}
\label{tbl:news_topics}
\end{table}
\begin{table}[t!]
\centering
\resizebox{\linewidth}{!}{%

\begin{tabular}{lrr}
\toprule
  Media outlet & \# of articles & Media-level ideology \\ 
\midrule
Washington Times& 100& Far Right\\
CNN (Online News)& 58& Far Left\\
New York Times (News)& 52 & Lean Left\\
HuffPost& 51 & Far Left\\ 
Washington Post& 45 & Lean Left\\
Politico& 44 & Lean Left\\
USA TODAY& 38 & Lean Left\\
NPR (Online News)& 32 & Center\\
Newsmax (News)& 32 & Far Right\\
Townhall& 23 & Lean Right \\
The Hill& 23 & Central\\
Reuters& 21 & Central\\
BBC News& 15 & Central\\
Fox News (Online News)& 13 & Lean Right\\
Breitbart News& 11 & Lean Right\\
National Review& 11 &  Lean Right\\
Vox& 10 & Far Left\\
The Guardian& 10& Lean Left\\
Reason& 7 & Far Right\\
Christian Science Monitor& 7 & Center\\
Washington Examiner& 2 & Far Right\\
TheBlaze.com& 2 & Center\\
Wall Street Journal (News)& 1& Center\\
Salon& 1& Far Left\\
\midrule
Total & 609 & - \\
\bottomrule
\end{tabular}
}
\caption{Number of articles collected from each source and corresponding media-level ideology based on AllSides label.}
\vspace{-15pt}
\label{tbl:media_outlets}
\end{table}

\section{Annotation Guideline}
\label{annotation_guideline}

\textbf{Step 1: Entity annotation:} First, read the entire article and list the entities as well as their corresponding types. 
\textit{Main entities} are the major subjects, objects, or participants involved in the main events described in the article. 
\textit{Salient entities} broadly refer to political or notable figures that appear in the news stories even if they are not the main ones, including public figures, celebrities, or other important people, events, or subjects. 
Entities are always nominals (i.e., nouns or noun phrases), with examples and corresponding types listed below.

\textbf{Entity Name} must be a span of a text (please copy-and-paste from the text and stay with the surface form). If multiple versions exist, please use the most formal and complete span in the article. For example, the entity name for ``Mayor Ben Zahn II'', ``Ben Zahn'', ``Zahn'' or ``he'' (when referring to the entity) should be ``Mayor Ben Zahn III''.

\textbf{Entity Types} are from the set of \{People, Organization, Place, Event, Religion, Topic, Other\}. 

\smallskip
\noindent\textbf{Step 2: Sentiment annotation:} Next, read one sentence at a time and for each sentence annotate the sentiment held by one mentioned entity (subject) towards another entity (object) in the triplet format of $<$subject, sentiment, object$>$. Either the subject, the object, or both must be in the entity list annotated in Step 1. Do not annotate relationships where neither subject nor object is in the entity list, but feel free to add to these list if you discover any entities you missed in step 1. If there are multiple triplets in a sentence, please annotate all fairly clearly sentiments in that sentence. If a sentence does not contain any triplet with at least one entity, leave it blank.

\textbf{Subject or Object values} are from \{entities, “Not in the list”, and “None”\}

\textbf{``Not in the list''} should be used when a subject or object does not appear in the Entity list from Step 1. Note that when neither subject nor object appear in the list, the triplet should not be coded. 

\textbf{``None''} is used in cases where the subject or the object does not exist, or can not be identified. 

\textbf{Sentiment values} are from the \{Positive, Negative\}.

Note: Subject and object must be 1) explicitly mentioned in the text or 2) implicitly referred to by a pronoun (e.g., he, his) or by their titles (e.g., US President => Joe Biden), or 3) can be straightforwardly inferred from the context, e.g., the speaker identity is mentioned in previous sentences.

\smallskip
\noindent\textbf{Step 3: Article-level entity-targeted sentiment annotation:} Next, having read the whole article, please annotate the overall article-level sentiments towards all listed entities based on your reading. If you are unsure about the sentiment, please mark it ``Unknown''.

\textbf{Article-level sentiment values} are \{Very Positive, Positive, Slightly Positive, Neutral, Slightly Negative, Negative, Very Negative, Unknown\}.

\smallskip
\noindent\textbf{Step 4: Entity ideology annotation:} Next, annotate the ideology of entities based on your reading. Entity ideologies must be determined or inferred based on a combination of your knowledge of the article, your knowledge of the overall political context, and your sentiment annotation. If there is no clear identifiable ideology associated with an entity, please mark it ``Not Applicable''.

\textbf{Entity ideology values} are from \{Very liberal, Liberal, Slightly liberal, Moderate, Slightly conservative, Conservative, Very conservative, Not Applicable\}. 

\smallskip
\noindent\textbf{Step 5: Media-source ideology annotation:} Finally, attempt to estimate the ideology of the media organization that published this article. If you are unsure about the ideology, please mark it ``Unknown''.

\textbf{Media-source ideology values} are from \{Very liberal, Liberal, Slightly liberal, Moderate, Slightly conservative, Conservative, Very conservative, Unknown\}. 

\smallskip
\noindent\textbf{Post-hoc conversion:} We further convert fine-grained labels obtained in steps 3 through step 5 to coarse-grained labels according to the nature of each task. For sentiment annotation, we convert them as 3-way labels. 
Specifically, we convert very positive and positive into one positive category, and similarly for very negative and negative. Then we merge slightly positive, neutral, and slightly negative into neutral. 
For ideological labels obtained in steps 4 and 5, in light of the 5-way annotation provided by AllSides, we also convert ours as 5-way labels by merging very liberal and liberal into liberal, and similarly for very conservative and conservative.

\section{Details of Our Model}
\label{model_details}
This section is supplementary to \cref{decoder} and \cref{training} in the main content, with more details about mathematical formulations and implementation details. Our framework takes as input a multi-sentence document, $\vectorbold{x}=\{x_1,\ldots,x_{k+1},\ldots,x_{k+t},\ldots,x_L\}$, where the target sentence is in $\vectorbold{x}$, i.e., $\vectorbold{\tilde{x}}=\{x_{k+1},\ldots,x_{k+t}\}$.
Our model first generates a semantic graph $G$ as described in \cref{graph_construction}. $\vectorbold{x}$ and $G$ are consumed by BART encoder \citep{Lewis2020BARTDS} and graph encoder (\cref{graph_encoder}) separately, producing token representation, $\mathbf{H_T} \in \mathbb{R}^{m \times L}$, and  node representations, $\mathbf{H_G} \in \mathbf{R}^{m \times N}$, where $N$ denotes the number of nodes in graph $G$. Finally, stance triplets are generated by our decoder using improved \textit{in-parallel attention and information
fusion mechanisms} (\cref{decoder}). Moreover, we inject
Wikipedia knowledge to support the identification
of relations between entities, especially to provide additional information which is not present in texts, e.g., party affiliations and geopolitical relations (\cref{wiki}).

\begin{table*}[t]
\centering
\resizebox{0.99\linewidth}{!}{%
\begin{tabular}{lcccccc}
\toprule
\multicolumn{1}{c}{\multirow{2}{*}{}} & \multicolumn{3}{c}{\textbf{Full}} & \multicolumn{3}{c}{\textbf{Aspect}} \\ 
\cmidrule(lr){2-4}
\cmidrule(lr){5-7}
 & Acc. & F1 & Acc.Any & src-s & s-tgt & src-tgt \\
 \midrule
 \multicolumn{7}{l}{\textbf{Baselines (No Graph)}}\\
Sentence (Sen) & 7.26$_{\pm0.07}$ & 10.35$_{\pm0.31}$ & 12.39$_{\pm0.35}$ & 36.66$_{\pm0.50}$ & 23.81$_{\pm0.40}$ & 14.88$_{\pm0.37}$ \\
Sen + Context (Ctx) & 9.66$_{\pm0.18}$ & 14.08$_{\pm0.20}$ & 16.87$_{\pm0.24}$ & 45.24$_{\pm0.82}$ & 27.79$_{\pm0.35}$ & 20.01$_{\pm0.42}$ \\
Sen + Ctx + Entities & 11.32$_{\pm0.26}$ & 16.00$_{\pm0.34}$ & 19.15$_{\pm0.41}$ & 47.43$_{\pm1.16}$ & 30.03$_{\pm0.45}$ & 23.18$_{\pm0.34}$ \\
\midrule
\multicolumn{7}{l}{\textbf{Pipeline Models (Ours)}}\\
Graph & 12.03$_{\pm0.58}$ & 15.77$_{\pm0.84}$ & 19.86$_{\pm0.83}$ & 46.60$_{\pm0.95}$ & 31.07$_{\pm0.65}$ & 23.19$_{\pm0.79}$ \\
\noalign{\vskip 0.3ex}
\hdashline
\noalign{\vskip 0.3ex}
\,\, +  Oracle Entities & 31.84$_{\pm0.58}$ & 35.58$_{\pm0.76}$ & 44.87$_{\pm0.69}$ & 66.33$_{\pm1.08}$ & 55.50$_{\pm0.42}$ & 53.82$_{\pm0.61}$ \\
\midrule
\multicolumn{7}{l}{\textbf{End-to-end Models (Ours)}}\\
Graph \citep[seq. attn.;][]{cao-wang-2021-controllable} & 12.97$_{\pm0.34}$ & 17.22$_{\pm0.38}$ & 20.62$_{\pm0.46}$ & 50.58$_{\pm1.19}$ & 32.45$_{\pm0.66}$ & 24.76$_{\pm0.60}$ \\
Graph  &  13.62$_{\pm0.23}$ & 18.12$_{\pm0.30}$ & 21.78$_{\pm0.37}$ & 51.01$_{\pm0.80}$ & 32.65$_{\pm0.41}$ & \textbf{26.08}$_{\pm0.43}$ \\
\,\, + Multitask & 13.34$_{\pm0.22}$ & 18.16$_{\pm0.69}$ & 21.77$_{\pm0.84}$ & \textbf{52.10}$_{\pm0.88}$ & 32.06$_{\pm0.84}$ & 26.07$_{\pm0.75}$ \\
\,\, + Wiki & \textbf{13.74}$_{\pm0.21}$ & \textbf{18.24}$_{\pm0.26}$ & \textbf{21.87}$_{\pm0.31}$ & 51.41$_{\pm0.55}$ & \textbf{32.69}$_{\pm0.69}$ & 25.94$_{\pm0.46}$\\
\bottomrule
\end{tabular}
}
\caption{
Results on \data for E2E stance detection task, and breakdown of accuracy scores by aspects (average of 5 runs). 
Best results without oracle entities are in \textbf{bold}. 
Our graph-augmented model with Wikipedia knowledge performs the best on 4 out of 6 metrics, indicating the effectiveness of encoding knowledge.}
\label{tbl:result_std}
\end{table*}

\begin{table}[t]
\centering
\resizebox{0.9\linewidth}{!}{%
\begin{tabular}{lrr}
\toprule
 & Accuracy & Macro F1 \\
\midrule
BART \citep{Lewis2020BARTDS} & 86.32$_{\pm0.71}$ & 77.53$_{\pm0.71}$ \\
POLITICS \citep{politics} & 86.33$_{\pm0.83}$ & 77.48$_{\pm1.24}$ \\
DSE2QA \citep{park-etal-2021-blames} & 87.78$_{\pm0.56}$ & \textbf{79.90}$_{\pm0.80}$ \\
\midrule
Our model & \textbf{87.79}$_{\pm0.37}$ & 79.01$_{\pm0.66}$ \\
\bottomrule
\end{tabular}
}
\caption{Results on stance-only prediction for specified pairwise entities (average of 5 runs).
Our model performs on par with state-of-the-art models in stance detection tasks (POLITICS and DSE2QA). Our model performs on par with SOTA discriminative models. }
\label{tbl:result_dse_std}
\end{table}

\subsection{Decoder}
We decode with our improved multi-source fused decoder, improved upon Transformer decoder \citep{Vaswani2017AttentionIA}, to enable reasoning over both information sources: text and graph.

The key difference between the vanilla Transformer decoder and ours is the \textit{in-parallel cross-attention layer} which allows better integration of knowledge encoded in the two heterogeneous sources. Concretely, the cross attention to the text is formulated as:
\begin{equation} 
\mathbf{z}_T = \textrm{LayerNorm}(\mathbf{z}+\textrm{Attn}(\mathbf{z},\mathbf{H}_T))
\end{equation}
\noindent where $\mathbf{z}$ denotes the output from the self-attention layer, $\mathbf{H_T}$ is the token representations out of the text encoder, and  $Attn$ denotes the cross-attention mechanism as in \citet{Vaswani2017AttentionIA}. We can compute $\mathbf{z_G}$ in a similar manner that attends node representations ($\mathbf{H_T}$) from the graph encoder. 
\begin{equation} 
\mathbf{z}_G = \textrm{LayerNorm}(\mathbf{z}+\textrm{Attn}(\mathbf{z},\mathbf{H}_G))
\end{equation}
\noindent where $\mathbf{z}$ denotes the  output from the same self-attention layer, $\mathbf{H_G}$ is the node representations out of the graph encoder.

Our \textit{information fusion}  module enables the information interaction between textual ($\mathbf{z_T}$) and graph ($\mathbf{z_G}$) hidden states, to obtain the fused representation, $\mathbf{z'}$. 
We implement the following two operations for information fusion: (1) addition, i.e., $\mathbf{z'} =  \mathbf{z}_T + \mathbf{z}_G$, and (2) gating mechanism between $\mathbf{z}_T$ and  $\mathbf{z}_G$ similar to \cite{ZhaoNDK18}, as formulated below: 
\begin{equation}
\mathbf{z_f} = \textrm{GELU}(\mathbf{W^f}[\mathbf{z_T};\mathbf{z_G}]+\mathbf{b_f})
\end{equation}
\begin{equation}
\mathbf{\lambda} = \textrm{sigmoid}(\mathbf{W^\lambda}[\mathbf{z_T};\mathbf{z_G}]+\mathbf{b_\lambda})
\end{equation}
\begin{equation}
\mathbf{z} = \boldsymbol{\lambda} \odot \mathbf{z_f} + (1-\boldsymbol{\lambda}) \odot \mathbf{z_T}
\end{equation}

\noindent  where $\odot$ is element-wise product, and $\mathbf{W}^{\ast}$ and $\mathbf{b}_{\ast}$  are learnable. $\boldsymbol\lambda$ here denotes the learnable gate vector. The selection of operation is decided by the downstream task. Specifically, in experiments we use addition for task A and gating mechanism for task B.

\subsection{Training Objectives}
We adopt the cross entropy (CE) training objective that minimizes the following loss for model training.
\begin{equation}
\mathcal{L}_{stance} = -\sum_{(\mathbf{x},\mathbf{y}) \in D} \log p(\mathbf{y}|\mathbf{x})
\end{equation}

\noindent where the reference $\mathbf{y}$ is a sequence of ground-truth stance triplet(s), sorted by their entities' first occurrences in the target sentence $\tilde{\mathbf x}$, and $D$ denotes the training set. 
$\mathbf{x}$ is formatted as \textit{``<\textrm{s}> [\textrm{preceding context}] <\textrm{s}> [\textrm{target text}] <\textrm{/s}> [\textrm{succeeding context}] <\textrm{/s}>''}, where \textit{[$\cdot$]} indicates placeholders. 
Optionally, extracted entities can be paired with document $\mathbf{x}$ and then fed into the text encoder, in the format of \textit{``$\mathbf{x}$ <s> <ENT> $E_1$ <ENT> $E_2$ $...$''}.
$\mathbf{y}$ is formatted as \textit{``<\textrm{ENT}> [\textrm{source}] <\textrm{ENT}> [\textrm{target}] <\textrm{STANCE}> [\textrm{stance}]''}, where \textit{<$\cdot$>} is a separator and \textit{[$\cdot$]} is a placeholder.

\smallskip
\noindent\textbf{Variant with Node Prediction.} 
In addition to modeling entity interactions in the graph, we enhance the model by adding an auxiliary objective to predict the node salience, i.e., whether the corresponding entity should appear in the stance triplets $\mathbf{y}$ to be generated. 
This is motivated by the observation that $G$ usually contains excessive entity nodes, only a small number of which are involved in sentiment expression in the target sentence. 
Specifically, for each \textit{entity node} ${E_i}$, we predict its salience, i.e., $\hat{s_i}$, by applying affine transformation over its representation $\mathbf{h}_G$, followed by a sigmoid function
\begin{equation}
\hat{\mathbf{s}} = \textrm{sigmoid}( \mathbf{u}\mathbf{H_G^E})
\end{equation}
\noindent where $\hat{\mathbf{s}} = \{\hat{s}_1, \ldots, \hat{s}_N\}$ is the collection of all entity nodes, $\mathbf{H_G^E}$ is a matrix of node representations for entity nodes out of the graph encoder, and $\mathbf{u}$ is learnable during training.

We adopt the weighted binary cross entropy (BCE) training objective to minimize the loss, $\mathcal{L}_{node}$, over all \textit{entity nodes}.
\begin{equation}
\begin{aligned}
\mathcal{L}_{node} =  -\sum_{s_i} & w*s_i\log(\hat{s}_i) + \\  & (1-s_i)\log(1-\hat{s}_i)
\end{aligned}
\end{equation}

\noindent where $w$ controls loss weights on positive samples, and $s_i$ denotes the occurrence of entity node $E_i$ in the ground-truth stance triplet $\mathbf{y}$.

Finally, when the node prediction module is enabled, the overall loss for the \textbf{multitask} learning setup is $\mathcal{L}_{multi} = \mathcal{L}_{stance} + \mathcal{L}_{node}$.

\begin{figure}
\small
\centering
\begin{tabular}{ p{70mm} }
\toprule
... In her seven-page opinion, Justice Sotomayor wrote that the Trump administration had become too quick to run to the Supreme Court after interim losses in the lower courts. ``Claiming one emergency after another, the government has recently sought stays in an unprecedented number of cases, demanding immediate attention and consuming limited court resources in each,'' she wrote. {\ul ``And with each successive application, of course, its cries of urgency ring increasingly hollow.''} ... \\
\texttt{[0] Sonia Sotomayor \colorbox{green!15}{NEG} Donald Trump} \\
\noalign{\vskip 0.3ex}
\hdashline[5pt/4pt]
\noalign{\vskip 0.3ex}
\textbf{Sent.:} $<$Someone$>$  \colorbox{green!15}{NEG} Affordable Care Act\\
\textbf{Sent. + Cxt.:}  Supreme Court of the United States \colorbox{green!15}{NEG} Donald Trump \\
\textbf{Sent. + Cxt. + Ent.:}  Donald Trump \colorbox{green!15}{NEG} $<$Someone$>$ \\
\textbf{Graph model} (ours):   Sonia Sotomayor \colorbox{green!15}{NEG} Donald Trump \\
\textbf{Graph model + Wiki} (ours): Sonia Sotomayor \colorbox{green!15}{NEG} Donald Trump  \\
\midrule
... {\ul The actions drew charges of racism because more than 200,000 Black Michiganders would have their votes disallowed by \textbf{such an action}.} Palmer's comment that she would be willing to certify results in Detroit's suburbs -  which experienced some of the same clerical errors that Detroit did - but not in Detroit, was seen as particularly outrageous. Both Palmer and Hartmann changed their votes on certification to ``yes'' Tuesday after strong criticism and heartfelt appeals from citizens participating in the board meeting over Zoom. But in the \textbf{affidavits} signed late Wednesday, both said they feel they made those votes under pressure and false pretenses and would now like to rescind their votes. ... \\
\texttt{[0] $<$Someone$>$ \colorbox{green!15}{NEG}  affidavit } \\
\noalign{\vskip 0.3ex}
\hdashline[5pt/4pt]
\noalign{\vskip 0.3ex}
\textbf{Sent.:} $<$Author$>$ \colorbox{green!15}{NEG} Barack Obama \\
\textbf{Sent. + Cxt.:}   $<$Someone$>$ \colorbox{green!15}{NEG}  Wayne County Board of Canvassers \\
\textbf{Sent. + Cxt. + Ent.:} $<$Someone$>$ \colorbox{green!15}{NEG} Wayne County, Michigan \\
\textbf{Graph model} (ours): $<$Someone$>$ \colorbox{green!15}{NEG} Wayne County, Michigan \\
\textbf{Graph model + Wiki} (ours): $<$Someone$>$ \colorbox{green!15}{NEG} Wayne County, Michigan \\
\bottomrule
\end{tabular}
\caption{Additional error analysis on two more test samples. {\ul The underlined sentence is the target sentence.} In the top example, both of our model are able to capture the correct stance triplet on this challenging sample while the baselines all fail, showing the power of graph modeling of entity interaction. In the bottom example, none of the five models get it right. 
More powerful models should be developed to encode broader context and knowledge. 
For example, event-level co-reference resolution seems to be imperative in order to understand the connection between \textit{affidavit} and the phrase ``such an action''.
} 
\vspace{-5mm}
\label{tbl:error_analysis_additional}
\end{figure}

\section{\data Annotation Quality Control}
\label{data_quality_control}
We hold meetings on a weekly basis and give annotators timely feedbacks to resolve annotation disagreements and iteratively improve annotation quality. We randomly sample $10\%$ news stories and have them annotated by multiple people. We first evaluate on the overlapping ratio between a pair of entity sets extracted by two different people. The overlapping ratio is $55.5$\% after cross-document entity resolution is conducted. Though the overlapping ratio is not high, we do find that entities captured by one but not both can help complement one another's annotation. Next, for sentiment annotation, we compute the agreement level by comparing two annotators' sentiment annotations on items that are annotated by both. We reach $97$\% agreement level, showing high quality of our \data.
Further, a simple unadjusted agreement between AllSides media-level ideology label and annotator's perception of the article's ideological leaning is 0.77 out of 1.0.\footnote{We consider the difference within one level as correct matching, e.g., \textit{Far Left (0)} and \textit{Lean Left (1)} are matched.}

\section{Reproducibility Checklist}

For all experiments, we use the Adam optimizer \cite{KingmaB14} with a learning rate of 1e-5 and fine-tune up to 15 epochs. The batch size of all baselines and our models are 4. The gradient is clipped when its norm exceeds 5. 
We select the best model for each method using the accumulated loss on the dev set. In decoding, the batch size is 1. We also enable learning rate decay with a patience of 200 steps. The early stop is also enabled with a patience of $1,600$ steps. For all these other hyperparameters, we keep the default values.

\end{document}